\def\year{2021}\relax
\documentclass[letterpaper]{article} 
\usepackage{aaai21}  
\usepackage{times}  
\usepackage{helvet} 
\usepackage{courier}  
\usepackage[hyphens]{url}  
\usepackage{graphicx} 
\usepackage{natbib}  
\usepackage{caption} 
\usepackage[utf8]{inputenc} 
\usepackage[T1]{fontenc}    
\usepackage{url}            
\usepackage{booktabs}       
\usepackage{amsfonts}       
\usepackage{nicefrac}       
\usepackage{microtype}      

\usepackage{amsmath}
\usepackage{amsthm}
\usepackage{amsfonts}
\usepackage{amssymb}
\usepackage{nccmath}

\usepackage{algorithm}
\usepackage{algorithmic}

\usepackage[short]{optidef}

\usepackage{graphicx}
\usepackage{listings}

\newtheorem{thm}{Theorem}

\newtheorem{assume}[thm]{Assumption}

\DeclareMathOperator*{\loss}{loss}
\DeclareMathOperator*{\premium}{premium}

\DeclareMathOperator*{\nrmse}{nrmse}
\DeclareMathOperator*{\mean}{mean}

\newcommand{\Brack}[1]{\left( #1 \right)}

\newcommand*{\skipInArxiv}[1]{}
\newcommand*{\skipInEightPage}[1]{}

\title{Fairness in Forecasting and Learning Linear Dynamical Systems}

\author{Quan Zhou,\textsuperscript{\rm 1}
Jakub Marecek,\textsuperscript{\rm 2}
Robert N. Shorten \textsuperscript{\rm 1,3}\\}
\affiliations{
\textsuperscript{\rm 1} University College Dublin, Ireland\\
\textsuperscript{\rm 2} Czech Technical University in Prague, the Czech Republic\\
\textsuperscript{\rm 3} Imperial College London, UK \\
quan.zhou@ucdconnect.ie, jakub.marecek@fel.cvut.cz, r.shorten@imperial.ac.uk}

\begin{document}
\maketitle


\begin{abstract}
In machine learning, training data often capture the behaviour of multiple subgroups of some underlying human population.
When the amounts of training data for the subgroups are not controlled carefully, under-representation bias arises.
We introduce two natural notions of subgroup fairness and instantaneous fairness to address such under-representation bias in time-series forecasting problems.
In particular, we consider the subgroup-fair and instant-fair learning of a linear dynamical system (LDS) from multiple trajectories of varying lengths, and the associated forecasting problems.
We provide globally convergent methods for the 
learning problems using hierarchies of convexifications of non-commutative polynomial optimisation problems. 
Our empirical results on a biased data set motivated by insurance applications and the well-known COMPAS data set demonstrate both the beneficial impact of fairness considerations on statistical performance and encouraging effects of exploiting sparsity on run time.
\end{abstract}


\section{Introduction}

The identification of vector autoregressive processes with hidden components
from time series of observations is a central problem across 
Machine Learning, Statistics, and Forecasting \cite{WestHarrison}.
This problem is also known 
as proper learning of linear dynamical systems (LDS) in System Identification \cite{Ljung1999}.
As a rather general approach to time-series analysis,
it has applications ranging from learning population-growth models in 
actuarial science and mathematical biology to 
functional analysis in neuroscience. 
Indeed, one encounters either 
 partially observable processes \cite{ASTROM1965Single}
 or questions of causality \cite{pearl2009causality} that can be 
 tied to proper learning of LDS  \cite{geiger2015causal}
 in almost any application domain.

A discrete-time model of a linear dynamical system $\mathcal{L}=(G,F,V,W)$ 
\cite{WestHarrison} suggests that the random variable 
$Y_t \in\mathbb{R}^{m}$ capturing the observed component (output, 
observations, measurements) evolves over time $t\geq 1$ according to:

\begin{eqnarray}
\phi_{t}  &=& G \phi_{t-1} + w_t, \label{equ:state-process} \\ 
Y_t &=& F' \phi_t + v_t, \label{equ:observation-process}
\end{eqnarray}

where 
$\phi_t \in \mathbb{R}^{n}$ is the  hidden component (state) and  
$G \in \mathbb{R}^{n\times n}$ and $F\in\mathbb{R}^{n\times m}$ 
are compatible system matrices. Random variables $w_t,v_t$ capture 
normally-distributed process noise and observation noise, 
 with zero means and covariance matrices $W \in\mathbb{R}^{n\times n}$ 
 and $V \in\mathbb{R}^{m\times m}$, respectively.
In this setting, proper learning refers to identifying the 
quadruple $(G,F,V,W)$ given the observations $\{Y_t\}_{t\in\mathbb{N}}$ 
of $\mathcal{L}$. This also allows for the estimation of subsequent observations, 
in the so-called ``prediction-error'' approach to improper learning \cite{Ljung1999}.

We consider a generalisation of the proper learning of LDS, where:
\begin{itemize}
    \item There are a number of individuals
    $p \in \mathcal{P}$ within a population.
    The population $\mathcal{P}$ is partitioned into subgroups indexed by $\mathcal{S}$. 
    
    
    \item For each subgroup $s\in\mathcal{S}$, there is a set $\mathcal{I}^{(s)}$ of trajectories of observations available and each trajectory $i \in \mathcal{I}^{(s)}$ has observations for periods $\mathcal{T}^{(i,s)}$, possibly of varying cardinality $|\mathcal{T}^{(i,s)}|$.

    \item Each subgroup $s \in\mathcal{S}$ is associated with a LDS, $\mathcal{L}^{(s)}$.
    For all $i \in \mathcal{I}^{(s)}$, $s\in\mathcal{S}$,
    the trajectory $\{Y_t\}^{(i,s)}$, 
    for $t\in \mathcal{T}^{(i,s)}$, is hence 
    generated by precisely one LDS $\mathcal{L}^{(s)}$. 
\end{itemize}

    

Note that for notations, the superscripts denote the trajectories and subgroups while subscripts indicates the periods.
In this setting, under-representation bias \cite[cf. Section 2.2]{blum2019recovering}, 
where the trajectories of observations from one (``disadvantaged'') 
subgroup are under-represented in the training data, harms both accuracy of the
classifier overall and fairness in the sense of varying accuracy across the subgroups.
This is particularly important, if the problem is constrained 
to be subgroup-blind, i.e., constrained to consider only a single LDS as a model. 
This is the case, when the use of attributes distinguishing each subgroup can be 
regarded as discriminatory (e.g., gender, race, cf. \cite{gajane2017formalizing}). 
Notice that such anti-discrimination measures are increasingly stipulated by the legal systems, e.g., within product or insurance pricing, where the sex of the applicant cannot be used, despite being known.

%



A natural notion of fairness in subgroup-blind learning of LDS involves 
estimating the system matrices or forecasting the next output of a single LDS 
that captures the overall behaviour across all subgroups, 
while taking into account the varying amounts of training data for the 
individual subgroups. 
To formalise this, suppose that we learn one LDS $\mathcal{L}$ from the 
multiple trajectories and we define a loss function that measures the loss
of accuracy for a certain observation $Y_t^{(i,s)}$, for $t\in \mathcal{T}^{(i,s)}$, 
$i \in \mathcal{I}^{(s)}$, $s\in\mathcal{S}$ when adopting the forecast $f_t$ 
for the overall population. For $t\in \mathcal{T}^{(i,s)}$, $i \in \mathcal{I}^{(s)}$, 
$s\in\mathcal{S}$, we have

\begin{equation}
    \loss^{(i,s)}(f_t):=||Y_t^{(i,s)}-f_t||. \label{equ:loss}
\end{equation}

Let $\mathcal{T}^+=\cup_{i\in\mathcal{I}^{(s)},s\in\mathcal{S}}\mathcal{T}^{(i,s)}$. 
We know that $f_t$ is feasible only when $t\in\mathcal{T}^+$. Note that since 
each trajectory is of varying length, it is possible that at certain 
triple $(t,i,s)$, there is no observation and $Y_t^{(i,s)}$, 
$\loss^{(i,s)}(f_t)$ become infeasible. 


We propose two  objective to address the under-representation bias, which extend group fairness \cite{feldman2015certifying} to time series:

\begin{enumerate}
    \item \textbf{Subgroup Fairness}. 
    The objective seeks to equalise, across all  
    subgroups, the sum of losses for the subgroup.
    Considering the number of trajectories in each subgroup and the 
    number of observations across the trajectories may differ, 
    we include $|\mathcal{I}^{(s)}|,|\mathcal{T}^{(i,s)}|$ as weights:
    
    \begin{equation}
        \min_{f_t,t\in\mathcal{T}^+} \max_{s\in\mathcal{S}} \left \{
        \frac{1}{|\mathcal{I}^{(s)}|}
        \sum_{i \in \mathcal{I}^{(s)}} \frac{1}{|\mathcal{T}^{(i,s)}|}
        \sum_{t\in \mathcal{T}^{(i,s)}} \loss^{(i,s)}(f_t) \right \} \label{equ:obj-Subgroup-Fair}
    \end{equation}
    
    \item \textbf{Instantaneous Fairness}. The objective seeks to equalise 
    the instantaneous loss, by minimising the 
    maximum of the losses across all subgroups and all times: 
    
    
    \begin{equation}
        \min_{f_t,t\in\mathcal{T}^+}
        \left \{ 
        \max_{t\in\mathcal{T}^{(i,s)},i\in\mathcal{I}^{(s)},s\in\mathcal{S}} \left \{ \loss^{(i,s)}(f_t) \right \} \right \} \label{equ:obj-Instant-Fair}
    \end{equation}

\end{enumerate}





Following \cite{zhou2020proper}, we also cast proper and improper learning of a linear dynamical system with such fairness considerations as a non-commutative polynomial optimisation problem (NCPOP), which can be solved efficiently using a globally-convergent hierarchy of semidefinite programming (SDP) relaxations.

\subsection{Related Work}

This presents an algorithmic approach to addressing the under-representation bias studied by \cite{blum2019recovering} and within the imbalanced learning literature \cite[e.g.]{he2013imbalanced,brabec2020model} and presents a step forward within the fairness in forecasting studied recently by \cite{gajane2017formalizing,chouldechova2017fair,NIPS20199603}, as  outlined in the excellent survey of   
\cite{chouldechova2020snapshot}. It follows much work on fairness in classification, e.g., \cite{zliobaite2015relation,hardt2016equality,kilbertus2017avoiding,kusner2017counterfactual,chouldechova2020snapshot,aghaei2019learning}. 
It is complemented by several recent studies involving dynamics and fairness \cite{mouzannar2019fair,paassen2019dynamic,jung2020fair}, albeit not involving \emph{learning} of dynamics.
It relies crucially on tools developed in non-commutative polynomial optimisation \cite{Pironio2010,wang2019tssos,wang2020chordal} and non-commutative algebra \cite{gelfand1943imbedding,segal1947irreducible,mccullough2001factorization,helton2002positive}, which have not seen much use in Statistics and Machine Learning, yet.  

\section{Motivation}
\label{sec:motivation}

\paragraph{Insurance Pricing}
\label{sec:insurancepricing}

Let us consider two motivating examples.
One important application arises in Actuarial Science. In the European Union, a directive (implementing the principle of equal treatment between men and women in the access to and supply of goods and services), bars insurers from using gender as a factor in justifying differences in individuals' premiums. 
In contrast, insurers 
in many other territories
classify insureds by gender, because females and males have different behavior patterns, which affects insurance payments. Take the annuity-benefit scheme for example. It is a well-known fact that females have a longer life expectancy than males \cite{huang2020effect}. The insurer will hence pay more to a female insured over the course of her lifetime, compared to a male insured, on averag  \cite{thiery2006fairness}. Because of the directive, a unisex mortality table needs to be used. As a result, male insureds receive less benefits, while paying the same premium in total as the female subgroup \cite{thiery2006fairness}. Consequently, male insureds might leave the annuity-benefit scheme (known as the adverse selection), which makes the unisex mortality table more challenging to use in the estimation of the life expectancy of the ``unisex'' population, where female insureds become the advantaged subgroup.

Consider a simple actuarial pricing model of annuity insurance. Insureds enter an annuity-benefit scheme at time $0$ and each insured can receive 1 euro in the end of each year for at most 10 years on the condition that it is still alive. Let $p_t$ denotes how many insureds left in the scheme in the end of the $t^{th}$ year. Suppose there are $p_0$ insureds in the beginning and the pricing interest rate is $i$ $(i\leq 1)$. The formula of calculating the pure premium is in \eqref{equ:premium}, thus summing up the present values of payment in each year and then divided by the number of insureds in the beginning.

\begin{equation}
    \premium := \frac{\sum_{t=1}^{10} p_t \times (1+i)^{-t} }{p_0} \label{equ:premium}
\end{equation}

The most important quality $p_t$ is derived from estimating insureds' life expectancy. 
Suppose the insureds can be divided into female subgroup and male subgroup and each subgroup only have one trajectory: $\{Y_t\}^{(\ \cdot \ ,f)}$ for female subgroup, $\{Y_t\}^{(\ \cdot \ ,m)}$ for male subgroup for $1\leq t\leq 10$, where the superscript $i$ is dropped. The two trajectories indicate how many female and male insureds are alive in the end of the $t^{th}$ year respectively. Both trajectories can be regarded as linear dynamic systems. We have

\begin{eqnarray}
    Y_t^{(\ \cdot \ ,f)} =& G^{(f)} Y_{t-1}^{(\ \cdot \ ,f)} + \omega_t^{(f)}, & 2\leq t\leq 10, \label{equ:female-LDS} \\
    Y_t^{(\ \cdot \ ,m)} =& G^{(m)} Y_{t-1}^{(\ \cdot \ ,m)} + \omega_t^{(m)}, & 2\leq t\leq 10, \label{equ:male-LDS} \\
    p_t =& Y_t^{(\ \cdot \ ,f)}+Y_t^{(\ \cdot \ ,m)}, & 1\leq t\leq 10, \label{equ:female+male}
\end{eqnarray}

where $\omega_t^{(f)}$ and $\omega_t^{(m)}$ are measurement noises while $G^{(f)}$ and $G^{(m)}$ are system matrices for female LDS $\mathcal{L}^{(f)}$ and male LDS $\mathcal{L}^{(m)}$ respectively. Note that these are state processes, without any observation process: the number of survivals can be precisely observed. 
To satisfy the directive, one needs to consider a unisex model:

\begin{eqnarray}
    f_t &=& G f_{t-1} + \omega_t, \, 2\leq t\leq 10,\label{equ:annuity-female+male-LDS}
\end{eqnarray}

where $2\leq t\leq 10$ and $\omega_t$ and $G$ pertain to the unisex insureds LDS $\mathcal{L}$. Subsequently, the loss functions for female (f) and male (m) subgroups are:

\begin{eqnarray}
\loss^{(\ \cdot \ ,f)}(f_t) :=& ||Y_t^{(\ \cdot \ ,f)}-f_t|| &,1\leq t\leq 10, \label{equ:loss-f-annuity}  \\
\loss^{(\ \cdot \ ,m)}(f_t) :=& ||Y_t^{(\ \cdot \ ,m)}-f_t|| &,1\leq t\leq 10, \label{equ:loss-m-annuity} 
\end{eqnarray}

Since the trajectories $\{Y_t\}^{(\ \cdot \ ,f)}$ and $\{Y_t\}^{(\ \cdot \ ,m)}$ have the same length and there is only one trajectory in each subgroup, the two objective \eqref{equ:obj-Subgroup-Fair}-\eqref{equ:obj-Instant-Fair} has the form:

\begin{equation}
        \min_{f_t,1\leq t\leq 10} \max \left \{ 
        \sum_{t=1}^{10} \loss^{(\ \cdot \ ,f)}(f_t),
        \sum_{t=1}^{10} \loss^{(\ \cdot \ ,m)}(f_t) \right \}
\end{equation}

\begin{equation}
        \min_{f_t,1\leq t\leq 10} \left \{
        \max_{1\leq t\leq 10, s\in\{f,m\}} \left \{
        \loss^{(\ \cdot \ ,s)}(f_t) \right \} \right \}
\end{equation}

\paragraph{Personalised Pricing}
\label{sec:Personalised Pricing}
Another application arises in personalised pricing (PP). 
For example, Amazon has been found \cite{OECDbackground} to sell certain products to regular consumers at higher prices. This is legal, albeit questionable.
In contrast, gender-based price discrimination 
\cite{abdou2019gender}
 violates \cite{OECDbackground}  anti-discrimination laws in many jurisdictions. 

Let us consider an idealised example of PP: Consider a soap retailer, whose customers contain female and male subgroups. Each gender has a specific dynamic system modelling its willing to pay (``demand price'' of each subgroup), while the retailer should set a ``unisex'' price. As in the discussion of insurance pricing, we consider subgroups $S=\{$female, male$\}$ and use superscripts $(f),(m)$ to distinguish the related quantities. Unlike in insurance pricing, the demand price of each customer is regarded as a single trajectory. More importantly, since customers might start buying the soap, quit buying the soap, or move to other substitutes at different time points, those trajectories of demand prices are assumed to be of varying lengths. For example, a customer starts to buy the soap at time $3$ but decides to buy hand wash instead from time $7$. 

Let us assume there are $|\mathcal{I}^{(f)}|$ female customers and $|\mathcal{I}^{(m)}|$ customers in the overall time window $\mathcal{T}^+$.
Let $Y_t^{(i,s)}$ denote the estimated demand price at time $t$ of the $i^{th}$ customer in subgroup $s$. These evolve as:

\begin{eqnarray}
    \phi_{t}^{f} =& G^{(f)} \phi_{t-1}^{(f)} + \omega_t^{(f)} &, t\in\mathcal{T}^+, \label{equ:person-female-state}\\
    Y_{t}^{(i,f)} =& F^{(f)'} \phi_{t}^{(f)} + \nu_t^{(i,f)} &, t\in\mathcal{T}^{(i,f)}, i\in\mathcal{I}^{(f)}, \label{equ:person-female-observation}\\
    \phi_{t}^{m} =& G^{(m)} \phi_{t-1}^{(m)} + \omega_t^{(m)} &, t\in\mathcal{T}^+, \label{equ:person-male-state}\\
    Y_{t}^{(i,m)} =& F^{(m)'} \phi_{t}^{(m)}+ \nu_t^{(i,m)} &,t\in\mathcal{T}^{(i,m)}, i\in\mathcal{I}^{(m)}.\label{equ:person-male-observation}
\end{eqnarray}

The unisex model for demand price considers the unisex state $m_t$, the unisex system matrices $G,F$, and unisex noises $\omega_t,\nu_t$:

\begin{eqnarray}
    m_t =& G m_{t-1} + \omega_t &, \, t\in\mathcal{T}^+, \label{equ:person-uni-state}\\
    f_t =& F' m_{t} + \nu_t &, \, t\in\mathcal{T}^+. \label{equ:person-uni-observation}
\end{eqnarray}


For $\loss^{(i,f)}(f_t) := ||Y_t^{(i,f)}-f_t|| ,t\in\mathcal{T}^{(i,f)}, i\in\mathcal{I}^{(f)}$ and $\loss^{(i,m)}(f_t) := ||Y_t^{(i,m)}-f_t||,t\in\mathcal{T}^{(i,m)}, i\in\mathcal{I}^{(m)}$, the two objective \eqref{equ:obj-Subgroup-Fair}-\eqref{equ:obj-Instant-Fair} has the form:

\begin{equation}
        \min_{f_t,t\in\mathcal{T}^+} 
        \max_{s\in\mathcal{S}} \left \{
        \frac{1}{|\mathcal{I}^{(s)}|}\sum_{i=1}^{\mathcal{I}^{(s)}} \frac{1}{|\mathcal{T}^{(i,s)}|}\sum_{t\in\mathcal{T}^{(i,s)}} \loss^{(i,s)}(f_t) \right \}
\end{equation}

\begin{equation}
        \min_{f_t,t\in\mathcal{T}^+} \left \{ \max_{t\in\mathcal{T}^{(i,s)},i\in\mathcal{I}_s,s\in\mathcal{S}} \left \{ \loss^{(i,s)}(f_t) \right \} \right \}
\end{equation}

\skipInEightPage{We also refer to \cite{dong2020protecting} for further work on protecting customers' interests in personalised pricing via fairness considerations.}

\section{Our Models}
\label{sec:models}

We assume that the underlying LDS $\mathcal{L}^{(s)}=(G^{(s)},F^{(s)},V^{(s)},W^{(s)})$ of each subgroup $s\in\mathcal{S}$ all have the form of  \eqref{equ:state-process}-\eqref{equ:observation-process}, while only one subgroup-blind LDS $\mathcal{L}$ can be learned and used for prediction. The following model in \eqref{NFF_1}-\eqref{NFF_2} can be used to describe the subgroup-blind state evolution directly. 

\begin{align}
          m_t & = G m_{t-1} + \omega_t, \label{NFF_1} \\
          f_{t} & = F' m_t + \nu_t. \label{NFF_2} 
\end{align}

for $t\in\mathcal{T}^+$, where $m_t$ represents the estimated subgroup-blind state and $\{f_t\}_{t\in\mathcal{T}^+}$ is the trajectory predicted by the subgroup-blind LDS $\mathcal{L}$.


The objectives \eqref{equ:obj-Subgroup-Fair} and \eqref{equ:obj-Instant-Fair}, subject to \eqref{NFF_1}-\eqref{NFF_2}, yield two operator-valued optimisation problems. 
Their inputs are $Y_{t}^{(i,s)},t\in\mathcal{T}^{(i,s)},i\in\mathcal{I}^{(s)},s\in\mathcal{S}$, i.e., the observations of multiple trajectories and the multiplier $\lambda$. The operator-valued decision variables $\mathcal{O}$ include operators proper $F, G$, vectors $m_t, \omega_t$, and scalars $f_t, \nu_t,$ and z. Notice that $t$ ranges over $t\in\mathcal{T}^+$, except for $m_t$, where $t\in\mathcal{T}^+\cup\{0\}$. The auxiliary scalar variable $z$ is used to reformulate "$\max$`` in the objective \eqref{equ:obj-Subgroup-Fair} or \eqref{equ:obj-Instant-Fair}.
Since the observation noise is assumed to be a sample of mean-zero  normally-distributed random variable, we 
add the sum of squares of $\nu_t$ to the objective with a multiplier $\lambda$, seeking a  solution with $\nu_t$ close to zero. 
Overall, the subgroup-fair and instant-fair formulations read:

\begin{mini} 
	  {\mathcal{O}}{z + \lambda \sum_{t\geq 1} \nu_t^2}{\label{min:FairA}}{} 
	  \textrm{Subgroup-Fair}
	  \addConstraint{z}{\geq
	  \frac{1}{|\mathcal{I}^{(s)}|}
        \sum_{i \in \mathcal{I}^{(s)}} \frac{1}{|\mathcal{T}^{(i,s)}|}
        \sum_{t\in \mathcal{T}^{(i,s)}} \loss^{(i,s)}(f_t)}
        {, s\in\mathcal{S}}
	  \addConstraint{m_t}{= G m_{t-1}+\omega_t}{, t\in\mathcal{T}^+}
	  \addConstraint{f_t}{= F' m_{t}+\nu_t}{, t\in\mathcal{T}^+.}
\end{mini}

\begin{mini} 
	  {\mathcal{O}}{z + \lambda \sum_{t\geq 1} \nu_t^2}{\label{min:FairB}}{} 
	  \textrm{Instant-Fair}
	  \addConstraint{z}{\geq
	  \loss^{(i,s)}(f_t)}{, t\in\mathcal{T}^{(i,s)},i\in\mathcal{I}^{(s)},s\in\mathcal{S}}
	  \addConstraint{m_t}{= G m_{t-1}+\omega_t}{, t\in\mathcal{T}^+}
	  \addConstraint{f_t}{= F' m_{t}+\nu_t}{, t\in\mathcal{T}^+.}
\end{mini}

For comparison, we use a traditional formulation that focuses on minimising the overall loss: 
\begin{mini}
	  {\mathcal{O}}{
	  \sum_{s \in \mathcal{S}} \quad
	  \sum_{i \in \mathcal{I}^{(s)}} 
	  \sum_{t\in \mathcal{T}^{(i,s)}} \loss^{(i,s)}(f_t)
	  + \lambda \sum_{t\geq 1} \nu_t^2}{\label{min:Unfair}}{} 	  \quad\textrm{Unfair}
	  \addConstraint{m_t}{= G m_{t-1}+\omega_t}{, t\in\mathcal{T}^+}
	  \addConstraint{f_t}{= F' m_{t}+\nu_t}{, t\in\mathcal{T}^+.}
\end{mini}


To state our main result, we need a technical assumption
related to the stability of the LDS, 
which suggests that the operator-valued decision variables (and hence estimates of states and observations) remain bounded.
Let us define the quadratic module, following \cite{Pironio2010}.  Let $Q=\{q_i\}$ be the set of polynomials determining the constraints. 
The positivity domain $\mathbf{S}_Q$ of $Q$ are tuples $X=(X_1,\ldots,X_n)$ of bounded operators
 on a Hilbert space $\mathcal{H}$ making all $q_i(X)$ positive semidefinite.
The quadratic module $M_Q$ is the set of  $\sum_if_i^{\dag}f_i+\sum_i\sum_j g_{ij}^{\dag}q_ig_{ij}$ 
where $f_i$ and $g_{ij}$ are polynomials from the same ring. 
As in \cite{Pironio2010}, we assume: 

\begin{assume}[Archimedean]
\label{Archimedean}
Quadratic module $M_Q$ of \eqref{min:FairA} is Archimedean, i.e., there exists a real constant $C$ such that $C^2-(X_1^{\dag}X_1+\cdots+X_{2n}^{\dag}X_{2n})\in M_Q$.
\end{assume}

Our main result shows that it is possible to recover the quadruple $(G,F,V,W)$ of the subgroup-blind $\mathcal{L}$ with guarantees of global convergence:

\begin{thm}
\label{T:covergence}
For any observable linear system $\mathcal{L}=(G,F,V,W)$, 
for any length $\mathcal{T}^+$ of a time window, 
and any error $\epsilon > 0$, 
under Assumption~\ref{Archimedean},
there is a convex optimisation problem from whose solution one can extract the best possible estimate of system matrices of a system $\mathcal{L}$ based on the observations, with fairness subgroup-fair considerations \eqref{min:FairA}, up to an error of at most $\epsilon$ in Frobenius norm. Furthermore, with suitably modified assumptions, the result holds also for the instant-fair considerations \eqref{min:FairB}.
\end{thm}


The proof is available in the full version of the paper on-line \cite{zhou2020fairness}.
It relies on the work of \cite{Pironio2010}, which shows the existence of a sequence 
of convex optimisation problems, whose objective function approaches the optimum of the non-commutative polynomial optimisation problem, and on the work of Gelfand, Naimark, and Segal \cite{gelfand1943imbedding,segal1947irreducible,klep2018minimizer}, which makes it possible to extract the minimiser of the non-commutative polynomial optimisation problem from the solution of the convex optimisation problem. 



\section{Numerical Illustrations}

\subsection{Generation of Biased Training Data}
\label{sec:Biased Training Data Generalisation}

To illustrate the impact of our models on data with varying degrees of under-representation bias, we consider a method for generating data resembling the motivating applications in Section~\ref{sec:motivation}, with varying degrees of the bias.
Suppose there is one advantaged subgroup and one disadvantaged subgroup, $S=\{$advantaged, disadvantaged$\}$ with trajectories $\mathcal{I}^{(a)}$ and $\mathcal{I}^{(d)}$ in each subgroup. Under-representation bias can enter training set in the following steps:

\begin{enumerate}
    \item Observations $Y^{(i,s)}_t$ are sampled from corresponding LDS $\mathcal{L}^{(s)}$. Thus each $Y^{(i,s)}_t\sim\mathcal{L}^{(s)}$.
    \item Discard some trajectories in $\mathcal{I}^{(d)}$, if necessary, such that $|\mathcal{I}^{(a)}|\geq|\mathcal{I}^{(d)}|$.
    \item Let $\mathcal{\beta}^{(s)},s\in\mathcal{S}$ denote the probability that an observation from subgroup $s$ stays in the training data and $0\leq \beta^{(s)}\leq 1$. Discard more observations of $\mathcal{I}^{(d)}$ than those of $\mathcal{I}^{(a)}$ so that $\mathcal{\beta}^{(a)}\geq \mathcal{\beta}^{(d)}$. If $\mathcal{\beta}^{(a)}$ is fixed at 1, the degree of under-representation bias can be controlled by simply adjusting $\mathcal{\beta}^{(d)}$.
    
\end{enumerate}
The last two steps discard more observations of the disadvantaged subgroup in the biased training data, so that  the advantaged subgroup becomes over-represented. 
Note that for small sample size, it is necessary to make sure there is at least one observation in each subgroup at each period. 



Consider that both subgroups $\mathcal{L}^{(s)},s\in\mathcal{S}$ have the same system matrices: $$G^{(s)}=\begin{bmatrix}
0.99 & 0 \\
1.0 & 0.2 \end{bmatrix},F^{(s)}=\begin{bmatrix}
1.1 \\ 0.8 
\end{bmatrix},$$
while the covariance matrices $V^{(s)},W^{(s)},s\in\mathcal{S}$ are sampled randomly from a uniform distribution over $[0,1)$ and $[0,0.1)$,  respectively. The initial states $m_0^{(s)}$ of each subgroups are $5$ and $7$. 
We set the time window to be $20$ across $3$ trajectories in the advantaged subgroup and $2$ in disadvantaged one, i.e., $|\mathcal{T}^+|=20$, $|\mathcal{I}^{(a)}|=3$ and $|\mathcal{I}^{(d)}|=2$. Then the bias is introduced according to the biased training data generalisation process described above, with random $\beta^{(s)},s\in\mathcal{S}$.

\begin{figure}[!t]
    \centering
    \begin{minipage}{.47\textwidth}
\includegraphics[width=0.95\textwidth,height=1.85in]{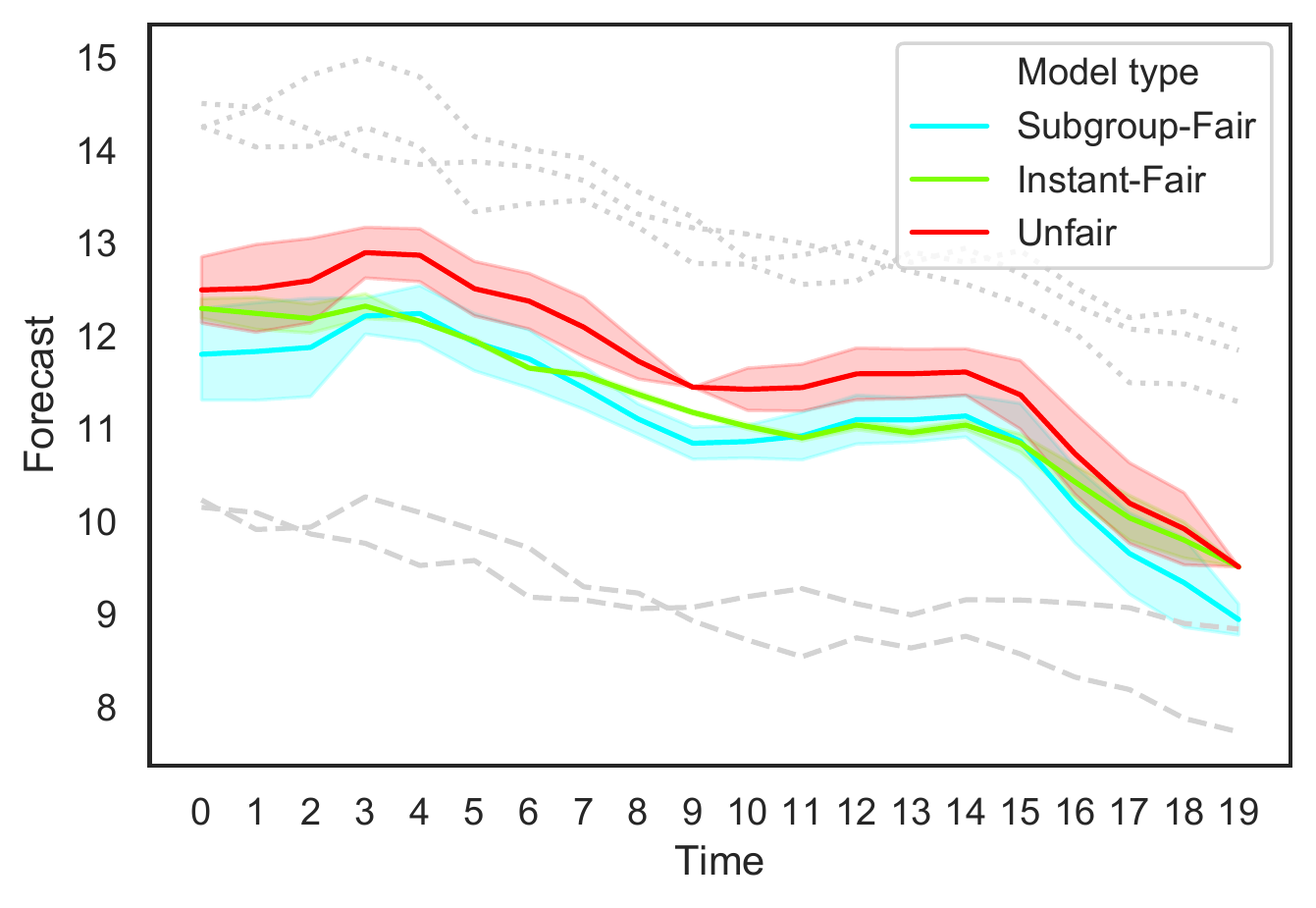} 
\caption{Forecast obtained using \eqref{min:FairA}-\eqref{min:Unfair}: the solid lines in primary colours with error bands display the mean and standard deviation of the forecasts over 10 experiments. For reference, dotted lines and dashed lines in grey denote the trajectories of observations of advantaged and disadvantaged subgroups, respectively, before discarding any observations.
}
\label{fig:LinePlot}
    \end{minipage}\hfill%
    \begin{minipage}{0.47\textwidth}
\includegraphics[width=0.99\textwidth]{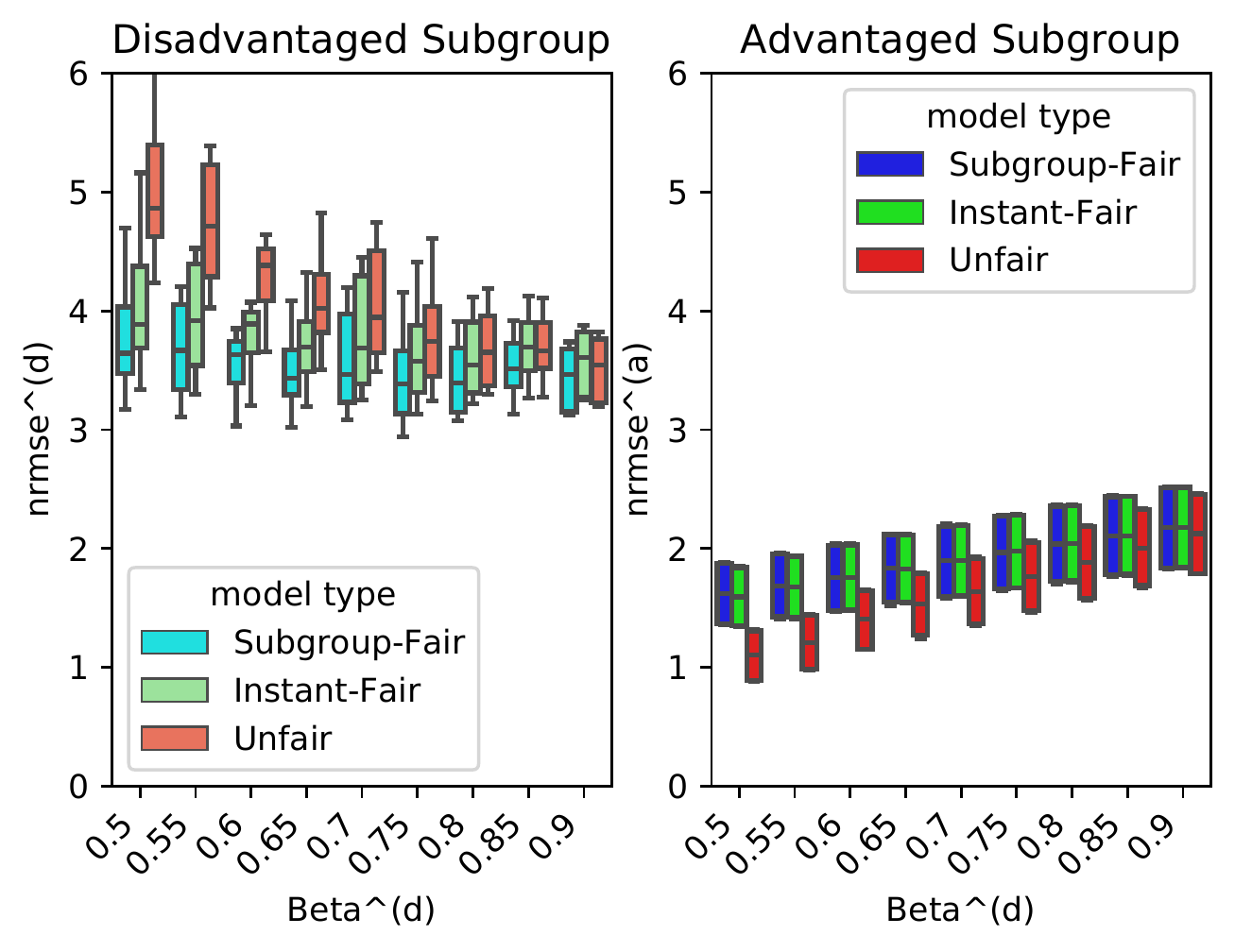}
\caption{Accuracy as a function of the degree of under-representation bias: The boxplot of $\nrmse^{(s)},s\in\mathcal{S}$ against $\mathcal{\beta}^{(d)}$, where $\mathcal{\beta}^{(d)}=[0.5,0.55,\dots,0.9]$, with boxes for the quartiles of $\nrmse^{(s)}$ obtained from $5$ experiments, using the observations in Figure~\ref{fig:LinePlot}.}
\label{fig:BetaPlot}
    \end{minipage}
\end{figure}

Figure~\ref{fig:LinePlot} shows the forecasts in 30 experiments on this example. For each experiment, the same set of observations $Y_t^{(i,s)},t\in \mathcal{T}^{(i,s)}$, $i \in \mathcal{I}^{(s)}$, $s\in\mathcal{S}$ is reused and the trajectories of advantaged and disadvantaged subgroups are denoted by dotted lines and dashed lines, respectively. However, the observations that are discarded vary across the experiments. Thus, a new biased training set is generated in each experiment, albeit based on the same ``ground set'' of observations. The three models \eqref{min:FairA}-\eqref{min:FairB} are applied in each experiment with $\lambda$ of 5 and 1, respectively, as chosen by iterating over integers 1 to 10. The mean of forecast $f$ and its standard deviation are displayed as the solid curves with error bands.

\subsection{Effects of Under-Representation Bias on Accuracy}

Figure~\ref{fig:BetaPlot} suggests how the degree of bias affects accuracy with and without considering fairness. 
With the number of trajectories in both subgroups set to 2, i.e. $|I_a|=|I_d|=2$ and $\mathcal{\beta}^{(a)}=1$, we vary the degree of bias  $\mathcal{\beta}^{(d)}$ within $0.5\leq\mathcal{\beta}^{(d)}\leq 1$.
To measure the effect of the degree on accuracy, we introduce the normalised root mean squared error (nrmse) fitness value for each subgroup $s\in\mathcal{S}$: 

\begin{equation}
    \mathrm{\nrmse^{(s)}}:=\sqrt{
    \frac{\sum_{i\in\mathcal{I}^{(s)}}\sum_{t\in\mathcal{T}^{(i,s)}}\Brack{ Y_t^{(i,s)}-f_t}^2}{\sum_{i\in\mathcal{I}^{(s)}}\sum_{t\in\mathcal{T}^{(i,s)}}\Brack{ Y_t^{(i,s)}-\mean^{(s)}}^2} },  \label{NRMSE}
\end{equation}

 where $\mean^{(s)}:=\frac{1}{|\mathcal{I}^{(s)}|} \sum_{i\in\mathcal{I}^{(s)}}\frac{1}{|\mathcal{T}^{(i,s)}|}\sum_{t\in\mathcal{T}^{(i,s)}} Y_t^{(i,s)}$. Higher $\nrmse^{(s)}$ indicates lower accuracy for subgroup $s$, i.e., the predicted trajectory of subgroup-blind $\mathcal{L}$ is further away from the subgroup. 

For the training data, the same set of observations $Y_t^{(i,s)},t\in \mathcal{T}^{(i,s)}$, $i \in \mathcal{I}^{(s)}$, $s\in\mathcal{S}$ in Figure~\ref{fig:LinePlot} is reused but $|I_a|=|I_d|=2$. Thus one trajectory in advantaged subgroup is discarded. Then, the biased training data generalisation process in Section~\ref{sec:Biased Training Data Generalisation} is applied in each experiment with $\mathcal{\beta}^{(a)}=1$ and the values for $\mathcal{\beta}^{(d)}$ selecting from $0.5$ to $0.9$ at the step of $0.05$.
At each value of $\mathcal{\beta}^{(d)}$, three models \eqref{min:FairA}-\eqref{min:Unfair} are run with new biased training data and the experiment is repeated for $5$ times. Hence, the quartiles of $\nrmse^{(s)}$ for each subgroup shown as boxes in Figure~\ref{fig:BetaPlot}.

One could expect that nrmse fitness values of advantaged subgroup in Figure~\ref{fig:BetaPlot} to be generally lower than those of the disadvantaged subgroup ($\nrmse^{(d)}\geq\nrmse^{(a)}$), leaving a gap. Those gaps narrow down as $\mathcal{\beta}^{(d)}$ increases, simply because more observations of disadvantaged subgroup remain in the training data. Compared the to ``Unfair'', models with fairness constraints, i.e., ``Subgroup-Fair'' and ``Instant-Fair'', show narrower gaps and higher fairness between two subgroups. More surprisingly, when $\nrmse^{(a)}$ decreases as $\mathcal{\beta}^{(d)}$ gets close to $0.5$, "Subgroup-Fair" model still can keep the $\nrmse^{(d)}$ at almost the same level, indicating a rise in overall accuracy. This is in contrast with results \cite{zliobaite2015relation,dutta2019information} from classification.

\begin{figure}[!t]
\centering
\begin{minipage}{0.47\textwidth}
\centering{
\includegraphics[width=0.99\textwidth]{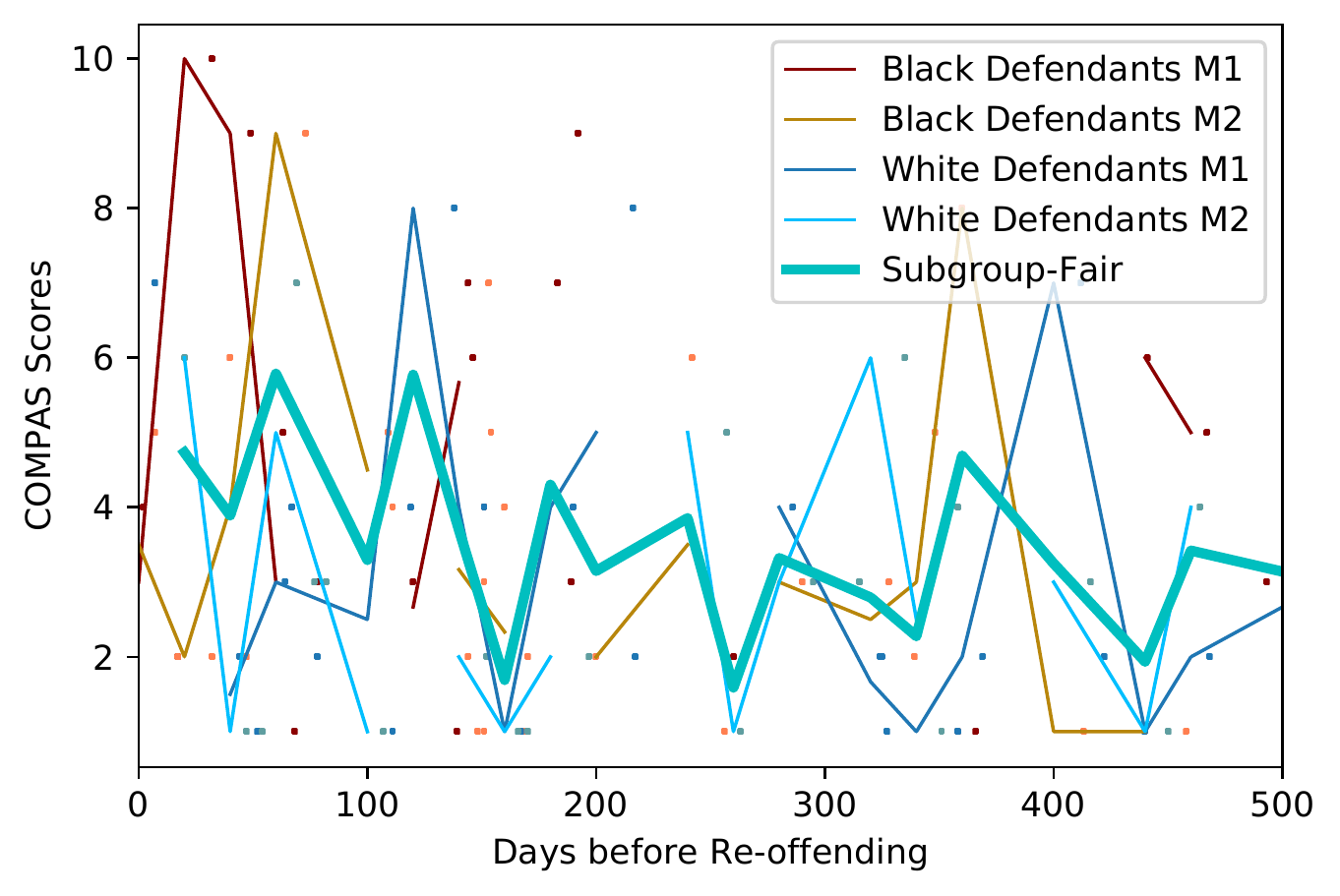}}
\caption{COMPAS recidivism scores of black and white defendants against the actual days before their re-offending. The sample of defendants' scores are divided into $4$ sub-samples based on race and type of re-offending, distinguished by colours. Dots and curves with the same colour denote the scores of one sub-sample and the trajectory extracted from the scores respectively. The cyan curve displays the result of "Subgroup-Fair" model with $4$ trajectories.}
\label{fig:COMPASPlot}
\end{minipage} \hfill%
\begin{minipage}{.47\textwidth}
\centering{
\includegraphics[width=0.99\textwidth]{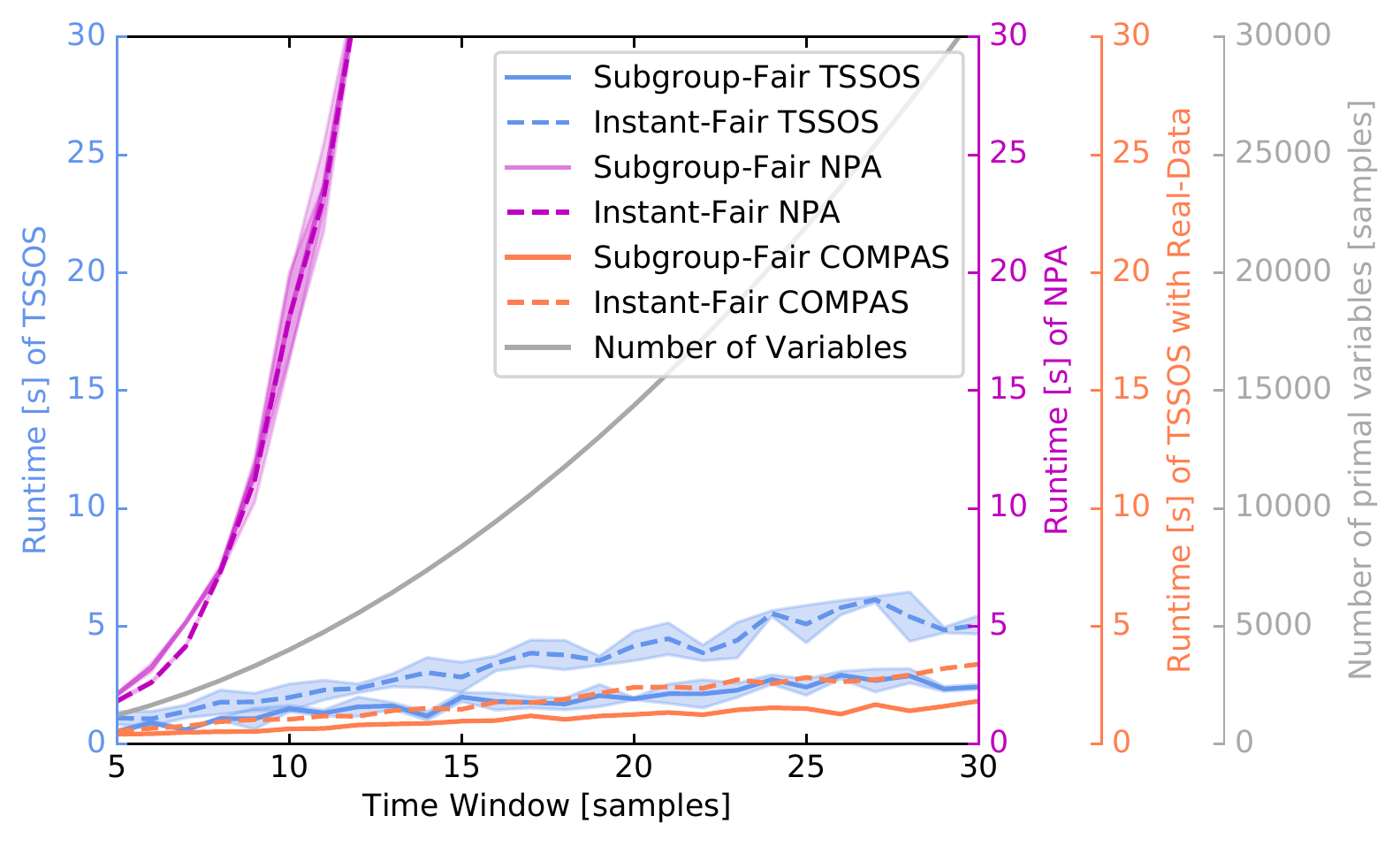} }
\caption{
The dimensions of relaxations and the run-time of SDPA thereupon as a function of the length of time window. Run-time of TSSOS and NPA is displayed in cornflower-blue and deep-pink curves, respectively, while the grey curve shows the number of variables in relaxations. Additionally, the run-time of the COMPAS dataset of Figure~\ref{fig:COMPASPlot} using TSSOS is also displayed as coral-coloured curves. For run-time, the mean and mean $\pm$ 1 standard deviations across $3$ runs are presented by curves with shaded error bands.
}
\label{fig:TimePlot}
\end{minipage} 
\end{figure}

\subsection{Run-Time}

Notice that minimising multivariate operator-valued polynomial optimization problems \eqref{min:FairA}-\eqref{min:Unfair} is non-trivial, but that there exists 
sparsity-exploiting variants (TSSOS)  of the globally convergent Navascués-Pironio-Acín (NPA) hierarchy used in the proof of Theorem \ref{T:covergence}.
See \cite{klep2019sparse,wang2019tssos,wang2020chordal,wang2020exploiting}.
The SDP of a given order in the respective hierarchy can be constructed using \texttt{ncpol2sdpa}\skipInEightPage{\footnote{\url{https://github.com/peterwittek/ncpol2sdpa}}} of  \cite{wittek2015algorithm} or the tools of \cite{wang2020exploiting} \skipInEightPage{\footnote{\url{https://github.com/wangjie212/TSSOS}}} and solved by \texttt{sdpa} of  \cite{yamashita2003implementation}. 
Our implementation is available on-line at \url{https://github.com/Quan-Zhou/Fairness-in-Learning-of-LDS}.


In Figure~\ref{fig:TimePlot}, we illustrate the run-time and size of the relaxations as a function of the length of the time window with the same data set as above (i.e., Figure~\ref{fig:LinePlot}). The grey curve displays the number of variables in the first-order SDP relaxation of "Subgroup-Fair" and "Instant-Fair" models against the length of time window. The deep-pink and cornflower-blue curves show the run-time of the first-order SDP relaxation of NPA and the second-order SDP relaxation of TSSOS hierarchy, respectively, on a laptop equipped by Intel Core i7 8550U at 1.80 Ghz. 
The results of "Subgroup-Fair" and "Instant-Fair" models are presented by solid and dashed curves, respectively. 
Since each experiment is repeated for three times, the mean and mean $\pm$ 1 standard deviation of run-time are presented by curves with shaded error bands. It is clear that the run-time of TSSOS exhibits a modest growth with the length of time window, while that of the plain-vanilla NPA hierarchy grows much faster.

\subsection{An Alternative Approach to COMPASS Dataset}

Finally, we wish to suggest the broader applicability of the two notions of subgroup fairness and instantaneous fairness. We use the well-known  dataset \cite{angwin2016machine} of estimates of the likelihood of recidivism made by the Correctional Offender Management Profiling for Alternative Sanctions (COMPAS), as used by courts in the United States. 
The dataset comprises of defendants' gender, race, age, charge degree, COMPAS recidivism scores, two-year recidivism label, as well as information on prior incidents. The two-year recidivism label denotes whether a person got rearrested within two years (label 1) or not (label 0). If the two-year recidivism label is $1$, there is also information of the recharge degree and the number of days until the person got rearrested.
We choose $119$ defendants with recidivism label 1, who are African-American or Caucasian, male, within the age range of 25-45, and with prior crime counts less than 2, with charge degree M and recharge degree M1 or M2.
The defendants are partitioned into two subgroups by their ethnicity and then partitioned by the type of their recharge degree (M1 or M2). Hence, we obtain the $4$ sub-samples.

In the days-to-reoffend-vs-score plot, such as Figure~\ref{fig:COMPASPlot}, dots suggest COMPAS recidivism scores of the $4$ sub-samples against the days before rearrest.
Curves capture models, either subgroup-dependent (plotted thin) or Subgroup-Fair (plotted thick).
The thick cyan curve is the race-blind prediction from our Subgroup-Fair method, which equalizes scores across the two subgroups.
Ideally, one should like to see a smooth, monotonically decreasing curves, overlapping across all subgroup-dependent models. For each sub-sample, aggregate deviation from the Subgroup-Fair curve would be similar to the aggregate deviations of other sub-samples. 

In the actual fact, dots are far removed from the ideal monotonically decreasing curve. Also, the subgroup-specific curves (plotted thin) are very different from each other (``subgroup-specific models are unfair''). Specifically, the red and yellow curves are above the sky blue and cornflower blue curves (``at the same risk level, white defendants get lower COMPAS scores''). 
Notice that the subgroup-dependent models are obtained as follows: we discretise time to $20$-day periods. 
For each subgroup, we check if anyone re-offends within $20$ days (the first period). If so, the (average) COMPAS score (for all cases within the 20 days) is recorded as the observation of first period of the trajectory of the sub-sample. If not, there is no observation of this period. We repeat this for the subsequent periods and for the three other sub-samples. 

In Figure~\ref{fig:TimePlot}, the coral-coloured curve (for the COMPAS dataset) suggests that the run-time remains modest, even as the length of the time window grows to 30. 


\section{Conclusions}
Overall, the two natural notions of fairness (subgroup fairness and instantaneous fairness), which we have introduced, may help establish the study of fairness in forecasting and learning of linear dynamical systems. We have presented globally convergent methods for the estimation considering the two notions of fairness using hierarchies of convexifications of non-commutative polynomial optimisation problems. The run-time of standard solvers for the convexifications is independent of the dimension of the hidden state.

An interesting direction for further research extends the two notions of fairness towards distributional robustness \cite{hashimoto18a} and uses the notions of fairness in constraints \cite{Donini2018}, as well as in the objective.

\section*{Acknowledgements}Quan's and Bob's work has been supported by the Science Foundation Ireland under Grant 16/IA/4610. Jakub acknowledges support of the OP RDE funded project CZ.02.1.01/0.0/0.0/16\_019/0000765 ``Research Center for Informatics''.

{
\fontsize{9.pt}{10.pt}
\selectfont
\bibliography{ref}
}


\clearpage
\section{Background}
\label{sec:background}
In this paper, we consider the case of multiple variants of the LDS and conduct proper learning of the LDS in a way of fairness using the technologies of non-commutative polynomial optimisation.
In Section~\ref{sec:background}, we firstly set our work in the context of system identification and control theory.
Secondly, we introduce the concept of fairness, which can be used to deal with multiple variants of the LDS.
In the end of this section, we provide a brief overview of 
non-commutative polynomial optimisation, pioneered by 
\cite{Pironio2010} and nicely surveyed by 
\cite{burgdorf2016optimization}, which is our key technical tool.
\subsection{Related work in system identification and control}
\label{sec:relwork}
Research within System Identification variously appears in venues associated with Control Theory, Statistics, and Machine learning. We refer to \cite{Ljung1999} and \cite{tangirala2014principles} for excellent overviews of the long history of research in the field, going back at least to \cite{ASTROM1965}. In this section, we focus on pointers to key more recent publications. 
In improper learning of LDS, a considerable progress has been made in the analysis of predictions for the expectation of the next measurement using auto-regressive (AR) processes. In \cite{anava13}, first guarantees were presented for auto-regressive moving-average (ARMA) processes. In \cite{arima_aaai}, these results were extended to a subset of autoregressive integrated moving average (ARIMA) processes.  \cite{Jakub} have shown that up to an arbitrarily small error given in advance, AR($s$) will perform as well as \emph{any} Kalman filter on any bounded sequence. 
This has been extended by \cite{tsiamis2020online} to Kalman filtering with logarithmic regret.
Another stream of work within improper learning focuses on sub-space methods \cite{katayama2006subspace,van} and spectral methods. 
\cite{tsiamis2019sample,tsiamis2019finite} presented the present-best guarantees for traditional sub-space methods.
Within spectral methods, \cite{hazan2017learning} and \cite{hazan2018spectral} have considered learning LDS with input, employing certain eigenvalue-decay estimates of Hankel matrices in the analyses of an auto-regressive process in a 
dimension increasing over time.
We stress that none of these approaches to improper learning  are ``prediction-error'': They do \emph{not} estimate the system matrices.

In proper learning of LDS, many state-of-the-art approaches consider the least-squares method, despite complications encountered in unstable systems \cite{faradonbeh2018finite}. \cite{simchowitz2018learning} have  
provided non-trivial guarantees for the ordinary least-squares (OLS) estimator 
in the case of stable $G$ and there being no hidden component, i.e., $F'$ being an identity and $Y_t = \phi_t$. 
Surprisingly, they have also shown that more unstable linear systems are easier to estimate than less unstable ones, in some sense. 
\cite{simchowitz2019learning} extended the results to allow for a certain pre-filtering procedure.
\cite{SarkarRakhlin} extended the results to cover stable,
marginally stable, and explosive regimes.

Our work could be seen as a continuation of the least squares method to processes with hidden components, with guarantees of global convergence.
In Computer Science, our work could be seen as an approximation scheme \cite{vazirani2013approximation},
as it allows for $\epsilon$ error for any $\epsilon > 0$.



\subsection{Fairness in machine learning}
\label{sec:fair}

The last two years have seen an unprecedented explosion in attention of fairness in the field of artificial intelligence and machine learning \cite{chouldechova2020snapshot}.

In machine learning, the training set might have biased representations of its subgroups, even when sampled with equal weight \cite{samadi2018price}.
Hence, a predictor trained from biased data set and aimed at maximising the overall accuracy might cause different distribution of errors in different subgroups, therefore not correspond
to a fair decision-making procedure. High accuracy need not be the primary goal of the system, especially when we consider that “accuracy” is measured on unfair data \cite{foulds2020intersectional}.
In facial recognition, \cite{buolamwini2018gender} find out that darker-skinned females are the most misclassified subgroup with error rates of up to 34.7\% while the maximum error rate for lighter-skinned males is 0.8\% as a result of the imbalanced gender and skin type distribution of the datasets of facial analysis benchmarks. Another threat facing us is gender bias shown in word embedding where the word \textit{female} is tender to be associated to \textit{receptionist} \cite{bolukbasi2016man}.

According to the clear summary in \cite{chouldechova2020snapshot}, the definition of fairness can be derived from a statistical notion and an individual notion. The statistical definition of fairness is to request a classifier’s statistic, such as raw positive classification rate (also sometimes known as statistical parity), false positive and false negative rates (also sometimes known as equalised
odds), be equalised across the subgroups so that the error caused by the algorithm be proportionately spread across subgroups \cite{chouldechova2020snapshot}. 
The statistical definition has a natural connection with Principal Component Analysis (PCA). Introduced in \cite{samadi2018price}, the Fair-PCA problem aims to minimise the maximum construction loss of different subgroups when looking for a lower dimensional representation. To solve the Fair-PCA problem, \cite{sharifi2019average} design an oracle-efficient algorithm while \cite{tantipongpipat2019multi} propose an algorithms based on extreme-point solutions of semi-definite programs. 

The individual definition asks for constraints that bind on specific pairs of individuals, rather than on a quantity that is averaged over groups \cite{chouldechova2020snapshot}, or in other words, it requires ``similar individuals should be
treated similarly'' \cite{dwork2012fairness}.
This definition is discussed less on account of its requirement of making significant assumptions even through it has strong individual level semantics that one's risk of being harmed by the error of the classifier are no higher than they are for anyone else \cite{sharifi2019average}.

We can introduce fairness to learning of LDS when dealing with multiple variants of the LDS. When estimating the next observation, one might be given several trajectories of observations from unknown variants of the LDS. In this case, fairness asks to find a suitable model that treats each LDS equally.

\subsection{Learning from Imbalanced data}

Traditional machines learning algorithms can be biased towards majority class over-prevalence \cite{chawla2003c4}, i.e., the under-representation bias \cite{blum2019recovering}. Also, the cost of misclassifying an abnormal event (minority class) as a normal event (majority class) is often relatively high \cite{chawla2008automatically,chawla2002smote}. For example, in the case of fraud, diseases, those cases are rare but able to cause serious damages, so it is of great interest to research. The benchmark of learning from imbalanced data was pioneered by \cite{chawla2002smote}. They proposed the Synthetic Minority Over-sampling Technique (SMOTE), such that a combination of over-sampling the minority class and under-sampling the majority class can efficiently improve the classifier performance.

The research in learning from imbalanced data has been extensively studied with a particular focus on classification and other predictive contexts as many real-world applications are already facing this problem \cite{torgo2017learning,fernandez2018smote}. SMOTE has been successfully extended to a variety of applications because of its simplicity and robustness \cite{cieslak2008learning}. Surprisingly, \cite{chawla2008automatically} provides an algorithm that automatically discovers the amount of re-sampling. 
One the other hand, \cite{moniz2017resampling} proposed the concept of temporal and relevance bias in extension of re-sampling strategies. For the clear journey of SMOTE, please refer to \cite{fernandez2018smote}.

Unlike the common solution of re-sampling, we address the under-representation bias from the view of optimisation, such that the ``loss'', or other statistical performance is equalised over majority and minority subgroups.

\subsection{Non-commutative polynomial optimisation}
\label{sec:ncpop}

In learning of the LDS, the key technical tool of this paper is non-commutative polynomial optimisation (NCPOP),
first introduced by \cite{Pironio2010}. 
Here, we provide a brief summary of their results, and refer to \cite{burgdorf2016optimization} for a book-length introduction.
NCPOP is an operator-valued optimisation problem with a standard form in Problem~\ref{NCPO}:

\begin{mini}
	  {(H,X,\phi)} {\langle\phi,p(X)\phi\rangle}{P:}{p*=}
	  \addConstraint{q_i(X)}{\succcurlyeq 0, }{i=1,\ldots,m}
	  \addConstraint{\langle\phi,\phi\rangle}{= 1,}{}\label{NCPO}
\end{mini}

where $X=(X_1,\ldots,X_n)$ is a bounded operator on a Hilbert space $\mathcal{H}$.
The normalised vector $\phi$, i.e., $\|\phi \|^2=1$ is also defined on $\mathcal{H}$ with inner product $\langle\phi,\phi\rangle$ equals to $1$. 
$p(X)$ and $q_i(X)$ are polynomials 
and $q_i(X)\succcurlyeq 0$ denotes that the operator $q_i(X)$ is positive semi-definite. 
Polynomials $p(X)$ and $q_i(X)$ of degrees $\deg(p)$ and $\deg(q_i)$, respectively,
can be written as:

\begin{equation}
    p(X)=\sum_{|\omega|\leq \deg(p)} p_{\omega} \omega,\quad
    q_i(X) = \sum_{|\mu|\leq \deg(q_i)} q_{i,\mu} \mu, 
    \label{LCoP}
\end{equation}

where $i = 1,\ldots,m$. 
Following \cite{akhiezer1962some}, we can define the moments on field $\mathbb{R}$ or $\mathbb{C}$, with a feasible solution $(H,X,\phi)$ of problem \eqref{NCPO}:

\begin{equation}
    y_{\omega} = \langle \phi, \omega(X) \phi \rangle, \label{DEFoMOMENT}
\end{equation}

for all $\omega \in \mathcal{W}_{\infty}$ and  $y_1=\langle \phi,\phi \rangle=1$.
Given a degree $k$, the moments whose degrees are less or equal to $k$ form a sequence of $y=(y_{\omega})_{|\omega| \leq 2k}$.
With a finite set of moments $y$ of degree $k$, we can define a corresponding $k^{th}$ order moment matrix $M_k(y)$:

\begin{equation}
    M_k(y)(\nu,\omega) = y_{\nu^{\dag}\omega} = \langle \phi, \nu^{\dag}(X)\omega(X) \phi \rangle,
    \label{equ:moment-matrix}
\end{equation}

for any $ |\nu|,|\omega| \leq k$ and a localising matrix $M_{k-d_i}(q_i y)$: 

\begin{align}
    M_{k-d_i}(q_iy)(\nu,\omega) & = \sum_{|\mu| \leq \deg(q_i)} q_{i,\mu} y_{\nu^{\dag}\mu\omega} \\ & = \sum_{|\mu| \leq \deg(q_i)} q_{i,\mu} \langle \phi, \nu^{\dag}(X) \mu(X) \omega(X) \phi \rangle, \notag
\end{align}

for any $|\nu|,|\omega| \leq k-d_i$, where $d_i=\lceil \deg(q_i)/2\rceil$. The upper bounds of $|\nu|$ and $|\omega|$ are lower than the that of moment matrix because $y_{\nu^{\dag}\mu \omega}$ is only defined on $\nu^{\dag}\mu\omega \in \mathcal{W}_{2k}$ while $\mu\in \mathcal{W}_{\deg(q_i)}$.

If $(H,X,\phi)$ is feasible, 
one can utilize the Sums of Squares theorem of \cite{helton2002positive} and \cite{mccullough2001factorization} to derive semidefinite programming (SDP) relaxations.
In particular, we can obtain a $k^{th}$ order SDP relaxation of the non-commutative polynomial optimization problem \eqref{NCPO} by choosing a degree $k$ that satisfies the condition of $2k\geq \max\{\deg(p),\max_i \deg(q_i)\}$.
The SDP relaxation of order $k$, which we denote $R_k$, has the form:

\begin{mini}
	  { y=(y_{\omega})_{|\omega|\leq 2k} }{\sum_{|\omega|\leq d} p_{\omega} y_{\omega}}{R_k:}{p^k=}
	  \addConstraint{M_k(X)}{\succcurlyeq 0}{}
	  \addConstraint{M_{k-d_i}(q_i X)}{\succcurlyeq 0, }{i=1,\ldots,m}
	  \addConstraint{\langle\phi,\phi\rangle}{= 1,}{}\label{NCPO-R}
\end{mini}

Let us define the quadratic module, following \cite{Pironio2010}.  Let $Q=\{q_i\}$ be the set of polynomials determining the constraints. 
The \emph{positivity domain} $\mathbf{S}_Q$ of $Q$ are tuples $X=(X_1,\ldots,X_n)$ of bounded operators
 on a Hilbert space $\mathcal{H}$ making all $q_i(X)$ positive semidefinite.
The \emph{quadratic module} $M_Q$ is the set of  $\sum_if_i^{\dag}f_i+\sum_i\sum_j g_{ij}^{\dag}q_ig_{ij}$ 
where $f_i$ and $g_{ij}$ are polynomials from the same ring. 
As in \cite{Pironio2010}, we assume:

\begin{assume}[Archimedean]
\label{ArchimedeanX}
Quadratic module $M_Q$ of \eqref{NCPO} is Archimedean, i.e., there exists a real constant $C$ such that $C^2-(X_1^{\dag}X_1+\cdots+X_{2n}^{\dag}X_{2n})\in M_Q$, where $X_{n+i},i\in\{1,\dots,n \}$ are defined to be $X_i^{\dag}$.
\end{assume}

If the Archimedean assumption is satisfied, \cite{Pironio2010} have shown that $\lim_{k \to \infty} p^k=p^*$ for a finite $k$.

\subsection{Proof of Theorem 2}

First, we need to show the existence of a sequence 
of convex optimisation problems, whose objective function approaches the optimum of the non-commutative polynomial optimisation problem.
As explained in the subsection above, 
\cite{Pironio2010} shows that, indeed, there are natural semidefinite programming problems, which satisfy this property.
In particular, the existence and convergence of the sequence is shown by Theorem 1 of \cite{Pironio2010}, which requires Assumption~\ref{Archimedean}.

Notice that we can use the so-called rank-loop condition of \cite{Pironio2010} to detect global optimality.
Once optimality is detected, it is possible to extract the global optimum $(H^*, X^*, \phi^*)$ from the optimal solution $y$ of problem $R_k$,
by Gram decomposition; cf. Theorem 2 in \cite{Pironio2010}.
Simpler procedures for the extraction have been considered, cf.  \cite{henrion2005detecting}, but remain less well understood. 

More broadly, we would like to show the extraction of the minimiser from the SDP relaxation of order $k(\epsilon)$ in the series is possible. 
There, one utilises the Gelfand--Naimark--Segal (GNS) construction \cite{gelfand1943imbedding,segal1947irreducible}, as explained in Section 2.2 of  \cite{klep2018minimizer},
which does not require the  rank-loop condition to be satisfied,
We refer to Section 2.2 of  \cite{klep2018minimizer} and Section 2.6 of  \cite{dixmier1969algebres} for details.

\end{document}